%% file: main.tex
\def\year{2017}
\pdfoutput=1
\documentclass[letterpaper]{article}

\usepackage{aaai17}
	
\usepackage{times}
\usepackage{helvet}
\usepackage{courier}
\usepackage{latexsym}
\usepackage{url}
\usepackage[draft]{hyperref}
\usepackage{amsmath,amssymb}
\DeclareMathOperator{\E}{\mathbb{E}}
\usepackage{floatrow}
\usepackage{graphicx}
\graphicspath{ {images/} }
\usepackage[font=small]{caption}
\usepackage{wrapfig}
\hypersetup{final}

\newcommand{\ParagraphHead}[1]{\noindent\textbf{#1}\hspace{5pt}}

\def\figref#1{Fig.~\ref{#1}}

\def\tabref#1{Table~\ref{#1}}

\usepackage[colorinlistoftodos]{todonotes} 
\nocopyright

\title{Contextual RNN-GANs for Abstract Reasoning Diagram Generation}

\author{
Arnab Ghosh\thanks{Equal Contribution}$^1$, Viveka Kulharia$^\ast$$^1$, Amitabha Mukerjee$^1$, Vinay Namboodiri$^1$, Mohit Bansal$^2$ \\
$^1$IIT Kanpur \quad $^2$UNC Chapel Hill \\ 
{\tt \{arnabghosh93,vivekakulharia\}@gmail.com,\{amit,vinaypn\}@iitk.ac.in}, \\ {\tt mbansal@cs.unc.edu} 
}

\begin{document}

\maketitle

\begin{abstract}
Understanding, predicting, and generating object motions and transformations is a core problem in artificial intelligence. Modeling sequences of evolving images may provide better representations and models of motion and may ultimately be used for forecasting, simulation, or video generation. Diagrammatic Abstract Reasoning is an avenue in which diagrams evolve in complex patterns and one needs to infer the underlying pattern sequence and generate the next image in the sequence. For this, we develop a novel Contextual Generative Adversarial Network based on Recurrent Neural Networks (Context-RNN-GANs), where both the generator and the discriminator modules are based on contextual history (modeled as RNNs) and the adversarial discriminator guides the generator to produce realistic images for the particular time step in the image sequence. We evaluate the Context-RNN-GAN model (and its variants) on a novel dataset of Diagrammatic Abstract Reasoning, where it performs competitively with 10th-grade human performance but there is still scope for interesting improvements as compared to college-grade human performance. We also evaluate our model on a standard video next-frame prediction task, achieving improved performance over comparable state-of-the-art.
\end{abstract}

\section{Introduction}

The recent success of machine learning and neural networks in Atari and Go~\cite{mnih2015human,silver2016mastering} has sparked a renewed interest in artificial intelligence models that can perform well at tasks which even humans find challenging. 
An important task in this category is abstract reasoning, which measures one's lateral thinking skills or fluid intelligence, i.e., the ability to quickly identify patterns, logical rules and trends in data, integrate this information, and apply it to solve new problems.
Specifically, we address the problem of diagrammatic abstract reasoning (DAR), a subset of Differential Aptitude Tests (DATs), which were introduced by \cite{bennett1947differential} to judge psychometric proficiency. It has featured in Intelligence Assessment Systems since 1950s and has been validated by \cite{berdie1951differential}, who showed that proficiency in DAT-DARs were predictors in engineering training. A DAR task involves the generation of a future diagram based on the sequential evolution of component patterns in the  given problem sequence.

\begin{figure}
	\centering
	\def\svgwidth{1\columnwidth}
	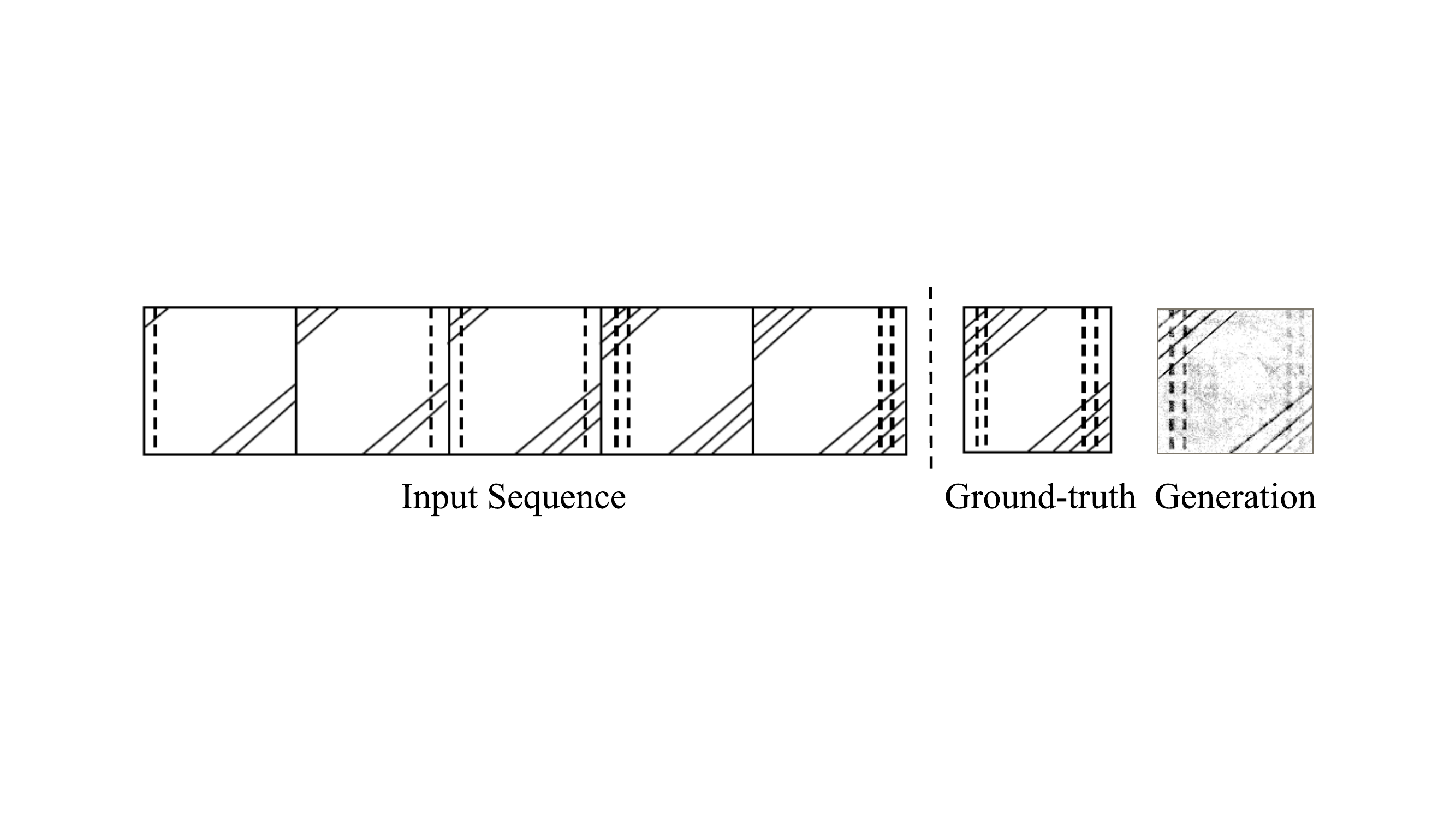
	\caption{Example abstract reasoning problem, where our model was able to generate an image very close to the correct answer. \vspace{-10pt}}
	\label{fig:admin4}
\end{figure}

\figref{fig:admin4} shows an example problem from our DAT-DAR dataset and highlights the intricacies of the reasoning involved in inferring the correct answer (i.e., the next image in the sequence). Different pattern components on both the sides and both the corners are changing in different and multiple ways, making it an interesting challenge to correctly generate the next image in the sequence.\footnote{An explanation of the ground truth is that the dashed line first goes to the left, then to the right, and then on both sides, and also changes from single to double, hence the ground truth should have double dashed lines on both the sides. On the corners, the number of slanted lines increase by one after every two images, hence the ground truth should have four slant lines on both the corners.}
Accurate generation models developed for such a reasoning task can be used for general AI applications such as forecasting and simulation generation. These models will also be useful for generation of real-world images and videos, a recent research direction in computer vision and deep learning~\cite{goodfellow2014generative,radford2015unsupervised,denton2015deep,im2016generating,mathieu2015deep}. In this direction, we present an application of our best models to the task of next frame generation in Moving MNIST videos.

Our sequential generation model is a temporally recurrent version of Generative Adversarial Networks (GANs)~\cite{goodfellow2014generative}, which we name Context-RNN-GANs\footnote{In Context-RNN-GAN, `context' refers to the adversary receiving previous images (modeled as an RNN) and the generator is also an RNN. The name distinguishes it from our simpler RNN-GAN model where the adversary is not contextual (as it only uses a single image) and only the generator is an RNN.}, where context refers to the sequential history (modeled as RNNs). This is especially well-suited to our sequential image-based reasoning tasks. In this model, the generator is an RNN which tries to generate the correct next image based on the previous sequence of images. The adversarial discriminator (playing  a minimax game with the generator) is also an RNN which gets the previous timesteps' images as context and the current timestep's image (either the one generated by the generator as a negative sample or the correct image from the dataset for that timestep as a positive sample).
In this way, the discriminator uses the images preceding the current timestep as contextual information to better distinguish whether the generator produced realistic images for the particular timestep in the sequence (as opposed to just producing a realistic image from the data distribution). 

We also develop novel image representations using an unsupervised Siamese Network~\cite{chopra2005learning} for modeling the joint representation of adjacent timesteps. This helps bring in much more information (as input features to the GAN model) about the temporal evolution across consecutive time steps. Modeling of the problem using an unsupervised setting and training as a sequential image language model on large quantities of sequences help the method to generalize better on unseen test sequences.

Empirically, we perform quantitative evaluation on the DAT-DAR dataset by first generating the successor image for each test sequence and then measuring similarity between the generated image and the candidate answer images, in a multiple-choice setting of the Intelligence Quotient (IQ) dataset. We then report the accuracy as the percentage of correct hits. We compare several baselines, model variants, and feature representations and find that our Context-RNN-GAN model with Siamese CNN features performs the best. Also, compared to human performance, we promisingly find that our final model is competitive with 10th-grade high school students but there is still scope for interesting improvements as compared to advanced engineering college students. We also demonstrate that our Context-RNN-GAN model can successfully model video next-frame generation via the Moving MNIST dataset, where we achieve improved performance over comparable state-of-the-art.

In summary, the main contributions of this paper are:
\begin{itemize}
\item A new abstract reasoning and image generation dataset with more than 1500 training problems (sequences of five images each, plus eight transformation types, leading to a total collection of more than 60000 training images); and an annotated evaluation test set of 100 problems.
\item A novel temporally contextual, RNN-based adversarial generation model, where the adversary has access to the full context of previous preceding images as context for deciding real vs fake sample for that timestep.
\item A novel feature representation of temporally adjacent images using a Siamese Network.
\item Strong performances on two datasets (DAR and MNIST videos), competitive with 10th-grade humans and comparable state-of-the-art.
\end{itemize}

\section{Related Work}

\cite{stern1914psychological} introduced IQ Tests to measure the success of an individual at adapting to a specific situation under a specific condition. Visual problems in intelligence tests have been among the earliest and continuously researched problems in AI. It has been looked at in terms of propositional logic beginning with \cite{evans1964program} and more recently by \cite{prade2011analogy}.
Recently, there has also been significant interest in building systems that compete with humans on a variety of tasks such as geometry-based problems~\cite{seo2015solving}, physics-based problems~\cite{novak1992uses}, repetition and symmetry detection~\cite{novak1992uses}, visual question answering~\cite{VQA}, and verbal reasoning and analogy~\cite{mikolov2013efficient,wang2015solving}.

Our task closely relates to the problem of Raven's Progressive Matrices. There, the problems are more constrained; one of the squares of the matrix is missing and the sequential pattern evolves along the columns and rows. This has been addressed using a propositional logic based framework~\cite{falkenhainer1989structure} and via a theorem prover based approach~\cite{bringsjord2004pulling}. However, we focus on a novel task which involves more unrestricted pattern movements, bigger datasets, and does not rely on representability in terms of propositional logic.

Our task is also closely related to the task of next frame prediction in videos~\cite{petrovic2006recursive} which involves predicting the next frame based on previous frames. However, the change across consecutive real-world video frames is extremely small as compared to the evolving shapes and changing spatial dynamics in our diagrammatic reasoning task. Hence, this poses several challenges to us different from the task of video next-frame generation, in which modeling optical flows plays a major role in producing better-looking next frames. Other related work in the video prediction direction include language modeling based approaches~\cite{ranzato2014video}, convolution-based LSTMs~\cite{patraucean2015spatio}, adversarial CNNs~\cite{mathieu2015deep}, context encoders ~\cite{pathakCVPR16context} and data-conditioned GANs~\cite{mirza2014conditional}.

Lastly, generative adversarial networks (GANs)~\cite{goodfellow2014generative} have also been extended with \emph{spatial} (and spatio-temporal) recurrence, attention, and structure~\cite{im2016generating,gregor2015draw,wang2016generative,vondrick2016generating}, whereas we specifically focus on \emph{temporal} recurrence constraints for frames of a video via RNNs. 
GANs have also been used for high resolution image generation~\cite{radford2015unsupervised}, image manipulation~\cite{zhu2016generative}, and text-to-image synthesis~\cite{reed2016generative}.

\cite{vondrick2016generating} generates videos from a single image using GANs but the discriminator judges the entire video rather than individual frames conditioned on previous frames. \cite{patraucean2015spatio} use Convolutional-LSTMs, similar to our GRU-RNN (with Shallow-CNN features) baseline but our final Context-RNN-GAN model with an adversarial loss gives better results. \cite{mathieu2015deep} predict multiscale videos using CNNs but only provide fixed number of previous frames as context to the discriminator which is not helpful for modeling short sequences. \cite{oh2015action} can generate long-term future frames in action dependent games but the transition of frames is mostly smooth with very similar consecutive frames, unlike our DAR task which involves discontinuous movements in an evolving diagrammatic pattern.

\section{Models}
\label{sec:models}
We first describe our primary Context-RNN-GAN model and then briefly discuss two simplifications of that model, namely RNN-GAN and a regular RNN. 

\begin{figure}[t]
    \centering
    \def\svgwidth{0.7\columnwidth}
    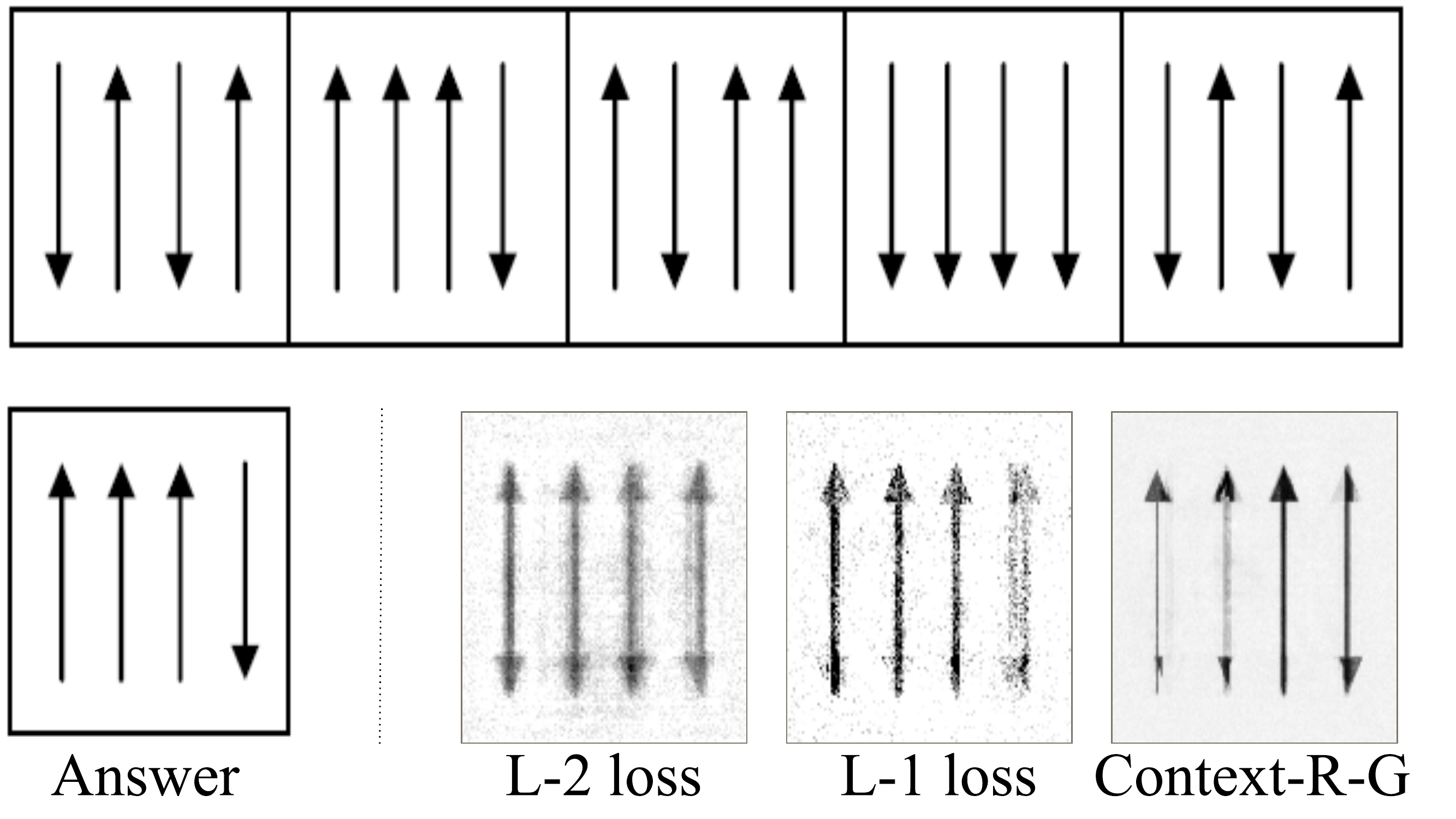
    \vspace{-4pt}
	\caption{Comparison of image generation quality for L-1 and L-2 loss functions (in GRU-RNN) vs our Context-RNN-GAN model. \vspace{-13pt}}
   \label{fig:visualization}
\end{figure}

The major motivation for using an adversarial loss was the shortcomings of L-2 and L-1 losses (\figref{fig:visualization}):
\begin{itemize}
	\item When using an L-2 loss function, some of the generated images were superimpositions of the component parts and were too cluttered.
	\item When using an L-1 loss function, although it was sharper than using an L-2 loss, it was missing some components of the actual diagrams.
\end{itemize}

\subsection{Context-RNN-GAN}
\begin{figure}[t]
    \centering
    \def\svgwidth{\columnwidth}
    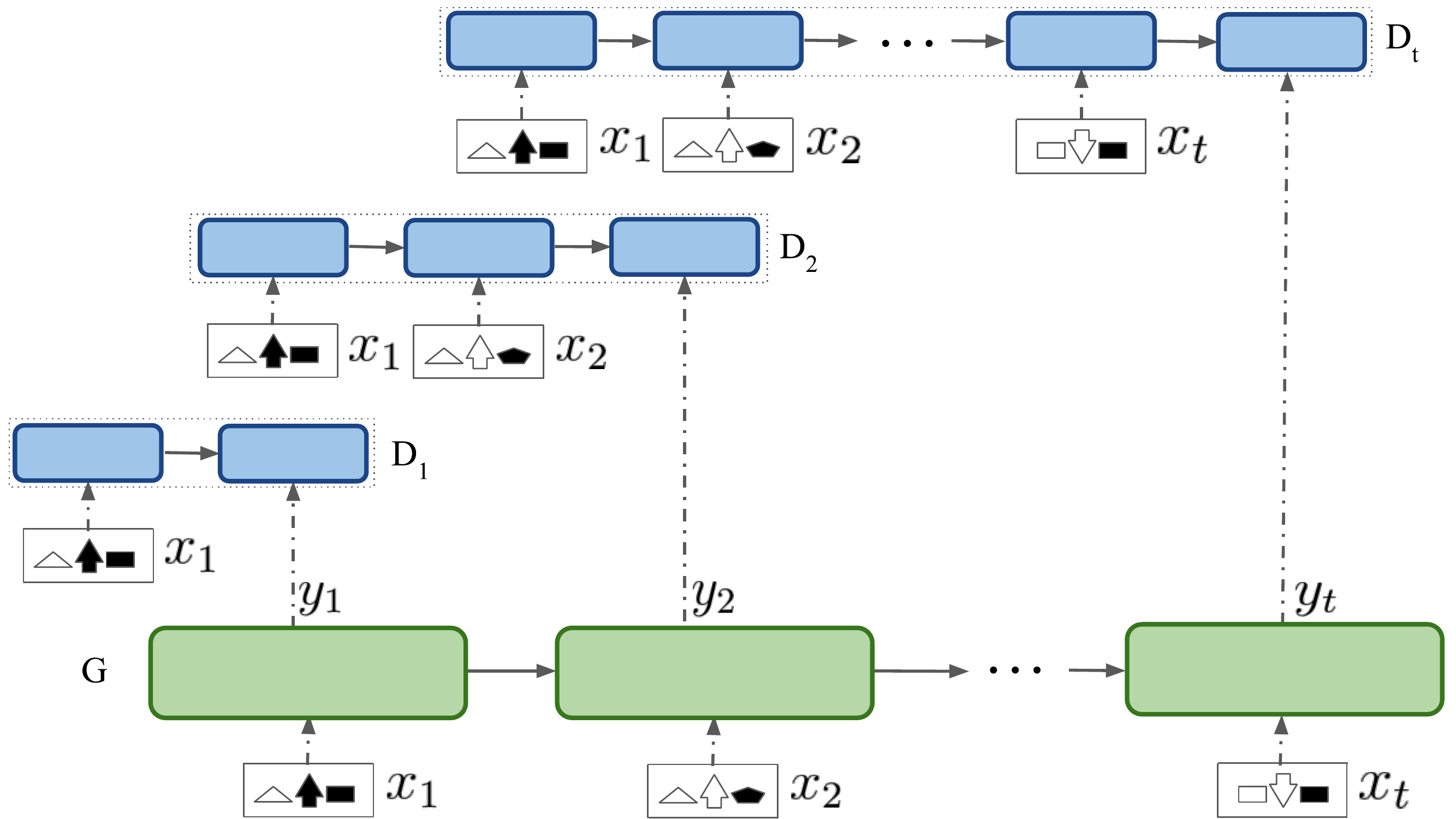
	\caption{Context-RNN-GAN model, where the generator G and the discriminator D (where $D_i$ represents its ith timestep snapshot) are both RNNs. G generates an image at every timestep while D receives all the preceding images as context to decide whether the current image output by G is real vs generated \emph{for that particular timestep}. $x_i$ are the input images. \vspace{-5pt}}
    \label{fig:contextrnn}
\end{figure}

Our Context-RNN-GAN model (\figref{fig:contextrnn}) uses the sequential structure of the diagrammatic abstract reasoning problem in a GAN framework, i.e., to generate the next image after a sequence of previous context images. 
The basic principle underlying our model is the same as that of the original GAN model by~\cite{goodfellow2014generative}, which is that the discriminator and generator play the following minimax game:

\begin{equation}
\label{eq3}
\begin{aligned}
	min_\beta max_\theta F\left(\beta,\theta\right) ={} 
	& \E_{x \sim p_{data}} [log p_\theta(y=1|x)] + \\
	&  \E_{x \sim p_\beta} [log p_\theta(y=0|x)]
\end{aligned}
\end{equation}

where $p_{data}$ is data's distribution, $p_\beta$ is generator's distribution with $\beta$ as the vector of parameters for the generator, and $p_\theta$ is discriminator's distribution with $\theta$ as the vector of parameters for the discriminator. The discriminator tries to distinguish between inputs $x$ sampled from the real data and from the generator's distribution by labeling them as 1 and 0 respectively. On the other hand, the generator tries to fool the discriminator by getting its generated image also labeled as 1 by the discriminator.
A GAN model is trained well when the discriminator cannot discriminate between the images generated by the generator and the images from the actual data distribution from which it is sampled.
Next, in our case, we importantly also add sequential context to both the generator and the discriminator of the GAN model (via RNNs) to capture the temporal sequence-of-images nature of our task, as described in detail next.

\paragraph{Generator}
We want the generator to generate an output image $y_{t}$ given the previous images (in the question) in sequential order $x_{1}$ to $x_{t}$, such that this generated image $y_t$ is as close as possible to the correct next timestep's image in the sequence $x_{t+1}$.
Therefore, we choose the generator to be a sequential RNN model, which generates the sequentially next image, $y_{t}=G(x_{1}\dots x_{t})$, trying to fool the discriminator into believing that these actually follow their preceding input, $x_{1}$ to $x_{t}$ (see \figref{fig:contextrnn}).
Note that our generator model can use LSTM-based or GRU-based or vanilla RNNs; we choose GRUs based on empirical evaluation.

\paragraph{Discriminator}
We want the discriminator to be able to decide whether a given image actually follows the previous sequence of images $x_1 \dots x_{t}$ (and not just whether the given image is from the real data distribution, unlike a traditional GAN model's discriminator). For this, we inject context into the discriminator too, by including the preceding (context) images, $x_1\dots x_{t}$, along with generator's generated image, $y_{t}$. It provides context for the discriminator to decide whether the generator's image actually follows the previous (context) images.
To model this context, we choose the discriminator to be a sequence model as well, namely a GRU-RNN. This sequence model is essentially a sequence-to-label encoding model which receives the context images as the preceding timesteps and the generator's image (or actual image) as the last timestep (see \figref{fig:contextrnn}). The final hidden state is mapped on to a sigmoid predicting whether it is an actual or fake image for that timestep.

\paragraph{Training the Discriminator (D)}

For a particular timestep (t+1), $x_{t+1}$ is the correct image for the timestep and $y_t=G(x_1 \dots x_t)$ is the generated image output by the generator G. We train the discriminator D such that $(x_1\dots x_t , x_{t+1})$ is classified as 1 (real) and $(x_1\dots x_t , y_{t})$ as 0 (fake).
Therefore the loss function we use to train D is:
\begin{equation}
\label{eq4}
\begin{aligned}
	L_{adv}^D ={} 
	& \sum_{t=1}^{timesteps-1} L_{bce} (D(x_1\dots x_t , x_{t+1}),1) \\
	&+ L_{bce}(D(x_1\dots x_t , y_{t}),0)\\
\end{aligned}
\end{equation}
where $L_{bce}$ is the binary cross entropy loss:
\begin{equation}
\label{eq5}
\begin{aligned}
L_{bce}(Y,\tilde{Y})={}
& - \tilde{Y} log (Y) + (1-\tilde{Y}) log(1-Y) \\
& \tilde{Y} \in \lbrace 0,1 \rbrace ,Y \in \left[0,1\right] \\
\end{aligned}
\end{equation}

\paragraph{Training the Generator (G)}
For a single sequence of images $ x_1 \dots x_{timesteps}$, keeping the weights of the discriminator D fixed, we minimize the adversarial loss:

\begin{equation}
\label{eq6}
L^G_{adv}=\sum_{t=1}^{timesteps-1} L_{bce}(D( x_1 \dots x_t , G( x_1 \dots x_t ) ),1)
\end{equation}

Minimizing the above loss implies that the G tries to adjust its weights so that the generated image  $y_t=G(x_1\dots x_t)$ is as close to a "real" image following $x_1 \dots x_t$ as judged by the current state of the discriminator. Just minimizing the above loss can lead to instability in training because the generator can generate images $y_t$ which are able to fool the discriminator but the manifold might be quite different to $x_{t+1}$ which it wants to model correctly.
Hence, similar to \cite{mathieu2015deep} and \cite{pathakCVPR16context}, we train the model with a  combination of $L^G_{adv}$ loss (adversarial loss) and $L_p$ loss, which is defined as:

\begin{equation}
\label{eq7}
L_p=\sum_{t=1}^{timesteps-1}   \lVert x_{t+1}-y_t \rVert_p^p
\end{equation}

The total loss to be minimized for the generator is then $ L^G = \lambda_{adv}L^G_{adv} + \lambda_{p} L_p $

\subsection{RNN-GAN}
RNN-GAN model is a simplified version of the Context-RNN-GAN model, where the discriminator is simply a Multilayer Perceptron (i.e. a Fully Connected Network) that only gets the generated (and real) images at the current timestep and no previous context image. The objective of this discriminator is to classify whether the image provided to it is an image from the dataset's distribution or is it generated by the generator. i.e. the discriminator is replaced from being $D(x_1 .. x_t , y_t)$ to being just $D(y_t)$.

\subsection{RNN}
Finally, the simplest model is a regular RNN (again with GRU units, similar to the generator modules above) which just models the sequence of images by trying to predict features of each image given the features of all the previous images in the sequence. We have tried it with $L_1$ and $L_2$ loss functions.

\subsection{Feed-forward Baseline}
The feed-forward network baseline is simply a fully-connected multi-layered perceptron with 2 layers, and is trained using the first 4 images $x_1 .. x_{t-1}$ to predict the fifth image $x_t$; and during testing, the last 4 images $x_2 .. x_{t}$ are used to predict the features of the answer image $x_{t+1}$.

\section{Feature Representations}
\label{sec:features}
In this section, we discuss the various image embedding methods that we used to create input features for the sequence models such as Context-RNN-GAN discussed above.

\ParagraphHead{Raw Pixels}
As the first baseline, each of the images was re-sized to the same dimension of 128$\times$128 and its raw pixel values were used as features, with row-wise stacking of the pixel values. Features generated by each model were re-sized to visualize the generation. 

\ParagraphHead{Histogram of Oriented Gradients}
HOG features~\cite{dalal2005histograms} are obtained by concatenating histograms of occurrences of gradient orientation in each of the cells of the images, hence capturing features corresponding to various edge types and proving to be a good feature detector.

\ParagraphHead{Autoencoder}
\cite{vincent2008extracting} have shown the effectiveness of autoencoders in reducing the dimensionality of the data. Hence, we used an unsupervised autoencoder with 1000 hidden units in the bottleneck layer and trained it on the images from the dataset. The activations of the bottleneck layer were then used as feature representations of the images.

\ParagraphHead{Pretrained CNN (OverFeat Network)}
We used 4096 features from the penultimate layers of an OverFeat network~\cite{sermanet2013overfeat} that won in the image localization task in ILSVRC 2013. Having been pre-trained on ImageNet dataset, it has been shown to be useful for tasks such as sketch recognition~\cite{yu2015sketch}.
The architecture consists of 7 stages, with 22 layers. Each stage consists of convolutions, rectified linear units (ReLU), and optionally of max-pooling layers. We extracted the output from the 21st layer that is the output of the fully connected layer in the seventh stage before final classification.

\ParagraphHead{Fine-tuned AlexNet model}
We fine-tuned an  AlexNet pre-trained on the Imagenet Challenge 2012~\cite{krizhevsky2012imagenet}. Labels were annotated for our DAT-DAR dataset's images by dividing each image into four quadrants and each of the four feature types (horizontal/vertical lines, slanted lines, and curved lines and the number of shaded regions) are counted in each quadrant of the image to get 16 labels for multioutput-multiclass classification. Counting of the features is done using the diagram parser used in \cite{seo2015solving}. The model used Euclidean loss and produced 16 outputs. 

\ParagraphHead{Shallow CNN}
We trained an end-to-end shallow CNN with only three convolutional layers (separated by ReLU and Pooling layers) followed by a ReLU, dropout and a fully connected layer. We trained it solely on our images' labels described above. The resultant features of the penultimate layer were used. The learning rate and the kernel size was similar to that used by~\cite{krizhevsky2012imagenet}. 

\ParagraphHead{Siamese CNN}
For the initial timesteps, regular sequential models such as LSTMs face difficulty in generating the next image in the sequence because of lack of sufficient context (because initial timesteps have lesser or no previous context). To resolve this, we propose to learn a joint embedding of (temporally) adjacent images and use that as features for our sequential RNN models.

For this, we created a Siamese network~\cite{chopra2005learning}, which uses the pair of shallow CNNs followed by a fully connected layer on top of each of the CNNs. The CNNs have parameters shared between them and they help find the distance between two images in a feature plane. The Siamese network tries to minimize the distance between similar images and maximize the distance between dissimilar images using contrastive loss~\cite{chopra2005learning}. Our Siamese network was trained by initializing the two shared CNNs with the parameters of the shallow CNN. It was then fine-tuned on our image dataset via a contrastive loss that uses every pair of two temporally adjacent images in a problem as similar examples and uses all other pairs of images from different problems as dissimilar examples. The activations of the two fully connected layers of the network are fed as features to our GAN models.

\section{Experimental Setup}

\ParagraphHead{Dataset}
We collected the data from several IQ test books and online resources \cite{aggarwalNonVerb}, \cite{sijwaliNonVerb}, \cite{guptaNonVerb} and \cite{prabhatNonVerb}.
We collected about 1500 training problems (15000 images) and an annotated test set of 100 problems, and then we used transforms such as rotation and mirror reflections across the axes to increase the  data by eight times, leading to a total of 12000 problem sequences to train on, each containing a sequence of 5 diagrams.   
Each test problem consists of five figures for the input sequence question and five as the answer choices, i.e., in a multiple-choice setup to allow easier quantitative evaluation. All the proposed models have been trained only on the five figures from the question part, i.e. the training set consisting of 12000x5 images was used, whereas the (sixth) correct answer figure was not used for training.
The models were validated (tuned) using the first 50\% of the answer figure set and tested on the remaining unseen 50\% of the answer figure set from the test/validation set of 100 questions.

\ParagraphHead{Training Details}
The best (validated) hyperparameters for the Context-RNN-GAN was a GRU with two layers of 400 hidden units each for the generator and a GRU with a single layer and 500 hidden units for the discriminator. 
The best (validated) hyperparameters for the RNN-GAN model was a GRU with two layers of 400 hidden units each for the generator and a mulilayered perceptron (MLP).
The final models of Context-RNN-GAN and the RNN-GAN models use $\lambda_{adv}=0.05$, $\lambda_{p}=1$, and they use $p=1$ for the DAT-DAR task and $p=2$ for the next-frame prediction task, similar to \cite{mathieu2015deep} and based on empirical evidence from experiments.
The best hyperparameters of the regular RNN was a GRU with one hidden layer of 1000 hidden units for both $L_1$ and $L_2$ based loss functions; adding more layers or more hidden units did not help. The best hyperparameters for the feedforward baseline was 2 layers with 1000 hidden units each for both $L_1$ and $L_2$ based loss functions and adding more layers or hidden units didn't help.
In all of the above models a dropout of 0.50 was applied for each hidden layer.
All of the models were trained using the Adam Optimizer \cite{kingma2014adam}.

\ParagraphHead{Evaluation Metrics}
In addition to qualitative evaluation via visualizations of the generated images, we importantly also perform quantitative evaluation by matching the generated image embedding with the embeddings of each of the five candidate answer images (based on cosine distance), and returning the closest matching image. The accuracy is then determined by the number of correct choices made by the model with respect to the total test set size.

\section{Results and Analysis}
\begin{figure*}[!ht]
    \def\svgwidth{1\columnwidth}
    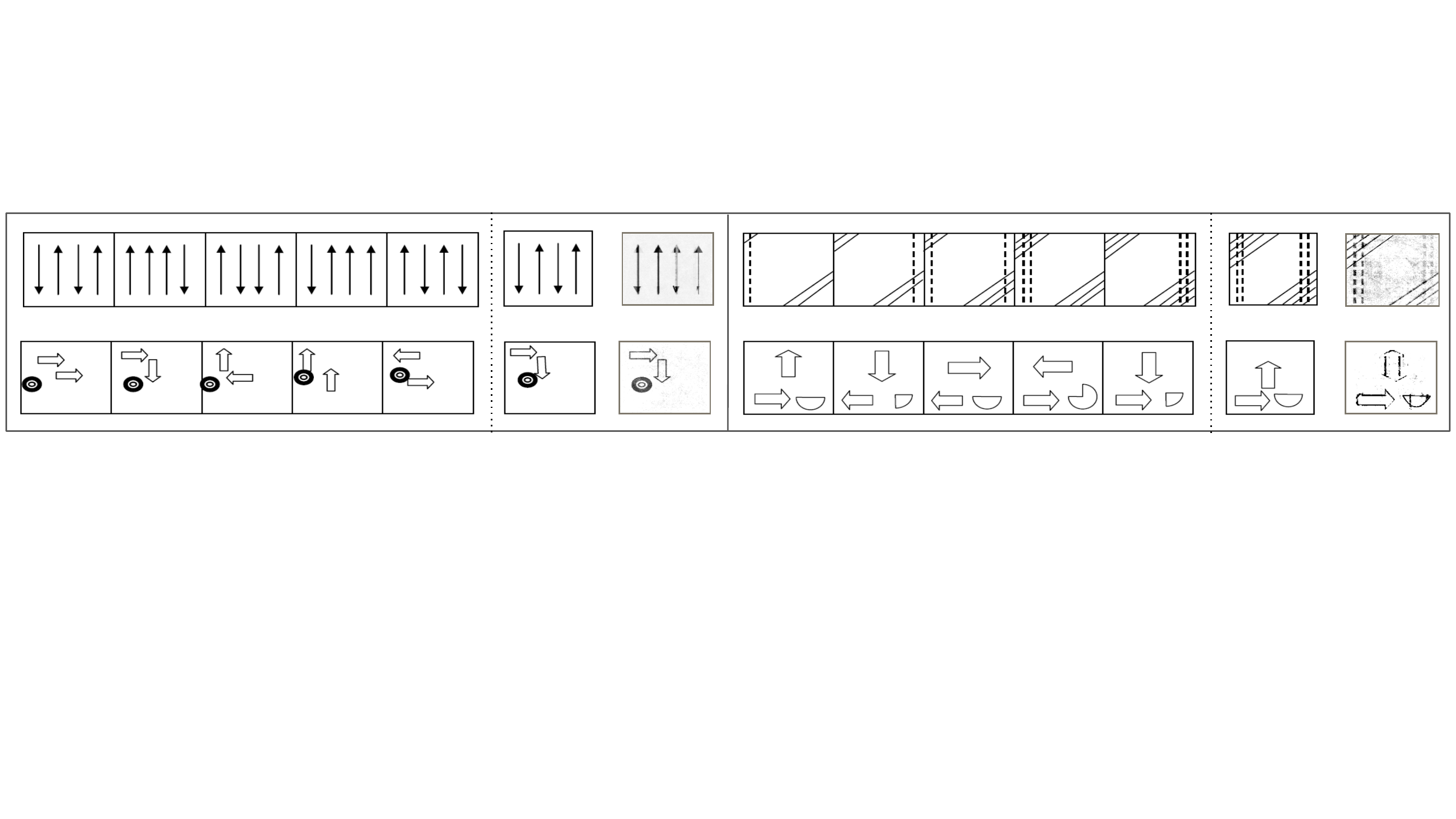
    \vspace{-17pt}
	\caption{Visualizations of image generation. In each of the four problems, the first five images are the question sequence, the second-last column is the ground-truth, and the last column is our model generation. \vspace{-6pt}}
   \label{fig:compare_visualization}
\end{figure*}

\subsection{Human Performance on DAR}

\begin{table}
	\centering
	\begin{tabular}{|c|c|c|}
		\hline 
	     & \ \ College-grade \ \ & \ \ 10th-grade \ \ \\ \hline
		 \ \ Age range \ \ & 20-22 & 14-16 \\ \hline         
		\#Students & 21 & 48 \\ \hline
		Mean &  44.17\% & 36.67\%  \\ \hline
		Std & 16.67\% & 17.67\% \\ \hline
		Max & 66.67\% & 75.00\% \\ \hline
		Min & 8.33\% & 8.33\% \\ \hline
	\end{tabular}
    \vspace{-6pt}
	\caption{Human performance on our DAR task. \vspace{-14pt}}
	\label{table:human}
\end{table}
\label{sec:human}

To test the proficiency of humans on our DAT-DAR dataset, we conducted a set of experiments on two sets of individuals. The problems were divided into sets of 12 problems each and they were given as much time as required to complete all the questions. They were also first given an example of a problem with an explanation of the answer.

\noindent\textbf{Advanced college students}: 21 senior students from the computer science department of a premier university took part in the first experiment.

\noindent\textbf{10th-grade high school students}: 48 students from the 10th grade of a reputed high school took part in the second experiment.

As can be seen in \tabref{table:human}, our diagrammatic abstract reasoning task is quite challenging even for humans, with the best performance of strong undergraduate college students being roughly $44\%$ and of 10th-grade students achieving $37\%$ accuracy. 

\subsection{Model Performance on DAR}
\label{sec:modelresults}

\begin{table}[t]
	\begin{small}
		\centering
		\begin{tabular}{|c|c|c|}
			\hline 
			\textbf{Features} &  \multicolumn{2}{|c|}{\textbf{Accuracy}}  \\
			\hline \hline 
			\multicolumn{3}{|c|}{RNN}\\
			\hline \hline 
			 & $L_2$ & $L_1$\\
			\hline \hline 
			Raw Pixels  & 26.0\% & 28.4\% \\
			HOG Features  & 28.0\% & 29.6 \% \\
			Autoencoder  & 11.6\% & 18.7 \% \\
            Image Labels & 21.1\% & 20.4 \% \\
			Pretrained CNN  & 18.6\% & 18.4 \%\\
			\ \ \ Fine-tuned AlexNet \ \ \ & \ \ \ 22.3\% \ \ \  & \ \ \ 20.3\% \ \ \  \\ 
			Shallow CNN & 28.6\% & 31.2 \%  \\
			Siamese CNN & 28.6\%  & 30.8 \%   \\
            \hline \hline
            \multicolumn{3}{|c|}{Feedforward Baseline}\\ 
            \hline \hline
            Raw Pixels & 20.8\% & 21.9\% \\
            HOG & 22.3 \% & 23.8 \% \\
            Shallow CNN & 24.2 \% & 23.6 \% \\
            Siamese CNN & 24.6 \% & 24.1 \% \\
			\hline \hline 
			\multicolumn{3}{|c|}{RNN-GAN}\\
			\hline \hline 
			Raw Pixels & \multicolumn{2}{|c|}{28.1\%}\\
			HOG & \multicolumn{2}{|c|}{30.1\%} \\ 
			Shallow CNN & \multicolumn{2}{|c|}{30.6\%} \\
			Siamese CNN \ \ & \multicolumn{2}{|c|}{31.2\%} \\ 
			\hline \hline 
			\multicolumn{3}{|c|}{Context-RNN-GAN}\\
			\hline \hline 
			\textbf{Raw Pixels} & \multicolumn{2}{|c|}{\textbf{30.3\%}} \\
			\textbf{HOG} & \multicolumn{2}{|c|}{\textbf{34.1}\%}  \\
			\textbf{Shallow CNN} & \multicolumn{2}{|c|}{\textbf{33.0\%}} \\
			\textbf{Siamese CNN} \ \  & \multicolumn{2}{|c|}{\textbf{35.4\%}} \\  \hline
		\end{tabular}
        \vspace{-4pt}
		\caption{Primary DAR results of Context RNN-GAN and its variants using several feature representations. \vspace{-4pt}}
		\label{table:1}
	\end{small}
\end{table}

\begin{table}[!ht]
	\begin{small}
		\centering
		\begin{tabular}{|c|c|c|}
			\hline
			\textbf{Model} & \textbf{CE} & \textbf{SE}\\
			\hline
			 LSTM & 341.2 & 208.1 \\
             AE-Conv-LSTM (w/o optical flow) & 262.6 & - \\
              \textbf{Context-RNN-GAN} & \textbf{241.8} & \textbf{167.9} \\
			 \hline
		\end{tabular}
	\end{small}
    \vspace{-4pt} 
	\caption{Results on Moving-MNIST videos. CE: cross-entropy error, SE: squared error. LSTM: Srivastava et al. (2015), AE-Conv-LSTM: Patraucean et al. (2015). \vspace{-10pt}}
	\label{table:2}
\end{table}

\begin{figure}[ht]
    \centering
    \def\svgwidth{\columnwidth}
    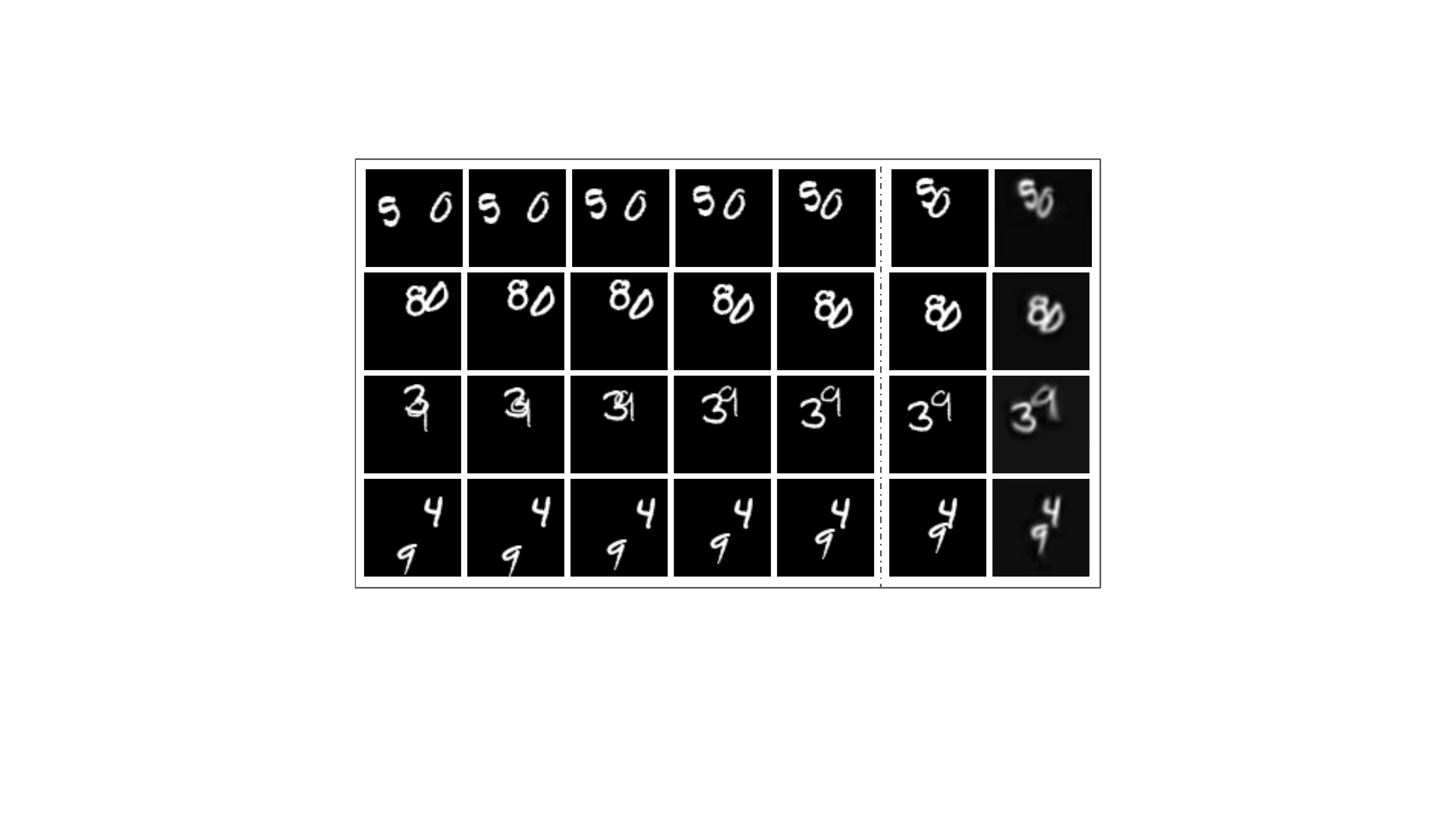
	\vspace{-18pt}
    \caption{Visualizations for Moving MNIST. First five images: question sequence, second-last column: ground-truth, last column: our model generation. \vspace{-8pt}}
   \label{fig:moving_MNIST}
\end{figure}

We next report the model results obtained via our proposed context-RNN-GAN model and its several simpler variants (specified in Models section), combined with the various feature representation methods (specified in Feature Representations section). In \tabref{table:1}, we first show baseline results for a regular RNN trained on different features. The shallow CNN, Siamese CNN, and HOG features perform best here. Next, we try these best feature settings (and the raw features) on our novel context-RNN-GAN model (and its simpler RNN-GAN variant). As shown in \tabref{table:1}, our primary context-RNN-GAN model, combined with our novel Siamese CNN features, obtains the best results, even competitive with the performance of 10th-grade human performance. However, there is still a lot of scope for interesting model and feature improvements compared to college-level human performance (and beyond), making this a new  challenging task and dataset for the community.

\paragraph{Next-Frame Generation on Moving-MNIST Videos}
We also tested our Context-RNN-GAN model on the popular Moving-MNIST task introduced by~\cite{srivastava2015unsupervised} and show results compared to their methods as well as the convolution-based LSTM model of~\cite{patraucean2015spatio}. For fair comparison, we report their results of the AE-Conv-LSTM model which does not model the optical flow. As shown in~\tabref{table:2}, we perform better than such comparable state-of-the-art of multiple metrics, CE (cross entropy error) and SE (squared error on image patches).

\subsection{Qualitiative Generation Visualization}
\label{sec:viz}

We also present quantitative evaluation via visualizations of the images generated by our model for both the DAR (\figref{fig:compare_visualization}) and the Moving-MNIST (\figref{fig:moving_MNIST}) tasks. We show cases where the generated image corresponds closely to the correct answer image. For example, in the DAR task (\figref{fig:compare_visualization}), the model is able to correctly infer and generate next-sequence diagrams with changing arrow directions, number and type of lines in different corners and sides, multiple shapes interacting in different ways, etc. Similarly, for the MNIST task (\figref{fig:moving_MNIST}), our model is able to correctly generate digits moving in different amounts and directions.

\section{Conclusion}

We presented a novel Context-RNN-GAN model that can generate images for sequential reasoning scenarios such as our task of diagrammatic abstract reasoning. When combined with useful feature representations such as those from Siamese CNNs, our model performs competitively with 10th-grade humans but there is still scope for interesting improvements as compared to college-level human performance, making this a novel challenging task for the generation community. Our sequential GAN model is also general enough to be useful for tasks such as next frame generations in a video (where we also achieve strong results on the Moving-MNIST dataset) and other similarly important AI tasks such as forecasting and simulation.

\section*{Acknowledgment} 
We thank the reviewers and Devi Parikh, Dhruv Batra (GaTech), Kundan Kumar (IIT Kanpur), Aravind Srinivas (IIT Madras), Deepak Pathak, Shubham Tulsiani (UC Berkeley), Greg Shakhnarovich (TTIC), and Shubham Tulsiani (UC Berkeley) for their helpful feedback.

{
\small
\small
\small
\small
\small
\bibliography{biblio}
}

\bibliographystyle{apalike}

\end{document}

%% file: best_visualization.pdf_tex
\begingroup%
  \makeatletter%
  \providecommand\color[2][]{%
    \errmessage{(Inkscape) Color is used for the text in Inkscape, but the package 'color.sty' is not loaded}%
    \renewcommand\color[2][]{}%
  }%
  \providecommand\transparent[1]{%
    \errmessage{(Inkscape) Transparency is used (non-zero) for the text in Inkscape, but the package 'transparent.sty' is not loaded}%
    \renewcommand\transparent[1]{}%
  }%
  \providecommand\rotatebox[2]{#2}%
  \ifx\svgwidth\undefined%
    \setlength{\unitlength}{768bp}%
    \ifx\svgscale\undefined%
      \relax%
    \else%
      \setlength{\unitlength}{\unitlength * \real{\svgscale}}%
    \fi%
  \else%
    \setlength{\unitlength}{\svgwidth}%
  \fi%
  \global\let\svgwidth\undefined%
  \global\let\svgscale\undefined%
  \makeatother%
  \begin{picture}(1,0.22)%
    \put(0,0){\includegraphics[clip,trim=2.5cm 5.6cm 2.5cm 5.5cm, width=\unitlength]{best_visualization.pdf}}
  \end{picture}%
\endgroup%

%% file: complete_context_RNN-GAN3.pdf_tex
\begingroup%
  \makeatletter%
  \providecommand\color[2][]{%
    \errmessage{(Inkscape) Color is used for the text in Inkscape, but the package 'color.sty' is not loaded}%
    \renewcommand\color[2][]{}%
  }%
  \providecommand\transparent[1]{%
    \errmessage{(Inkscape) Transparency is used (non-zero) for the text in Inkscape, but the package 'transparent.sty' is not loaded}%
    \renewcommand\transparent[1]{}%
  }%
  \providecommand\rotatebox[2]{#2}%
  \ifx\svgwidth\undefined%
    \setlength{\unitlength}{768bp}%
    \ifx\svgscale\undefined%
      \relax%
    \else%
      \setlength{\unitlength}{\unitlength * \real{\svgscale}}%
    \fi%
  \else%
    \setlength{\unitlength}{\svgwidth}%
  \fi%
  \global\let\svgwidth\undefined%
  \global\let\svgscale\undefined%
  \makeatother%
  \begin{picture}(2,0.5625)%
    \put(0,0){\includegraphics[width=\textwidth]{complete_context_RNN-GAN3.pdf}}%
  \end{picture}%
\endgroup%

%% file: compare_output.pdf_tex
\begingroup%
  \makeatletter%
  \providecommand\color[2][]{%
    \errmessage{(Inkscape) Color is used for the text in Inkscape, but the package 'color.sty' is not loaded}%
    \renewcommand\color[2][]{}%
  }%
  \providecommand\transparent[1]{%
    \errmessage{(Inkscape) Transparency is used (non-zero) for the text in Inkscape, but the package 'transparent.sty' is not loaded}%
    \renewcommand\transparent[1]{}%
  }%
  \providecommand\rotatebox[2]{#2}%
  \ifx\svgwidth\undefined%
    \setlength{\unitlength}{768bp}%
    \ifx\svgscale\undefined%
      \relax%
    \else%
      \setlength{\unitlength}{\unitlength * \real{\svgscale}}%
    \fi%
  \else%
    \setlength{\unitlength}{\svgwidth}%
  \fi%
  \global\let\svgwidth\undefined%
  \global\let\svgscale\undefined%
  \makeatother%
  \begin{picture}(1,0.5625)%
    \put(0,0){\includegraphics[width=\unitlength]{compare_output.pdf}}%
  \end{picture}%
\endgroup%

%% file: visualizations.pdf_tex
\begingroup%
  \makeatletter%
  \providecommand\color[2][]{%
    \errmessage{(Inkscape) Color is used for the text in Inkscape, but the package 'color.sty' is not loaded}%
    \renewcommand\color[2][]{}%
  }%
  \providecommand\transparent[1]{%
    \errmessage{(Inkscape) Transparency is used (non-zero) for the text in Inkscape, but the package 'transparent.sty' is not loaded}%
    \renewcommand\transparent[1]{}%
  }%
  \providecommand\rotatebox[2]{#2}%
  \ifx\svgwidth\undefined%
    \setlength{\unitlength}{768bp}%
    \ifx\svgscale\undefined%
      \relax%
    \else%
      \setlength{\unitlength}{\unitlength * \real{\svgscale}}%
    \fi%
  \else%
    \setlength{\unitlength}{\svgwidth}%
  \fi%
  \global\let\svgwidth\undefined%
  \global\let\svgscale\undefined%
  \makeatother%
  \begin{picture}(1,0.15)%
    \put(0,0){\includegraphics[clip, trim=0cm 7cm 0cm 4cm, width=\textwidth]{visualizations.pdf}}%
  \end{picture}
\endgroup%

%% file: moving_mnist.pdf_tex
\begingroup%
  \makeatletter%
  \providecommand\color[2][]{%
    \errmessage{(Inkscape) Color is used for the text in Inkscape, but the package 'color.sty' is not loaded}%
    \renewcommand\color[2][]{}%
  }%
  \providecommand\transparent[1]{%
    \errmessage{(Inkscape) Transparency is used (non-zero) for the text in Inkscape, but the package 'transparent.sty' is not loaded}%
    \renewcommand\transparent[1]{}%
  }%
  \providecommand\rotatebox[2]{#2}%
  \ifx\svgwidth\undefined%
    \setlength{\unitlength}{768bp}%
    \ifx\svgscale\undefined%
      \relax%
    \else%
      \setlength{\unitlength}{\unitlength * \real{\svgscale}}%
    \fi%
  \else%
    \setlength{\unitlength}{\svgwidth}%
  \fi%
  \global\let\svgwidth\undefined%
  \global\let\svgscale\undefined%
  \makeatother%
  \begin{picture}(1,0.5625)%
    \put(0,0){\includegraphics[clip,trim=6.6cm 4.1cm 6.6cm 3cm , width=\unitlength]{moving_mnist.pdf}}
  \end{picture}%
\endgroup%